%% file: active_learning.tex
\documentclass[letterpaper, 10 pt, conference]{ieeeconf}  

\IEEEoverridecommandlockouts

\usepackage{graphicx}
\usepackage{amsmath}
\usepackage{subcaption}
\usepackage{amssymb}
\usepackage{mlmacros}
\usepackage{float}
\usepackage{verbatim}
\usepackage{booktabs}
\usepackage{stfloats}
\usepackage{balance}

\usepackage{bm} 

\variables[obs,latent]{x,z}
\probdists{p,q}

\newcommand{\elbo}{\mathcal{L}_{\text{ELBO}}}

\newcommand{\z}{\mathbf{z}}

\newcommand{\x}{\mathbf{x}}
\newcommand{\J}{\mathbf{J}}
\newcommand{\G}{\mathbf{G}}

\newcommand{\cx}{f}
\newcommand{\cz}{\gamma}

\DeclareMathOperator{\MF}{MF}

\usepackage{todonotes}

\graphicspath{{figures/}}

\DeclareMathOperator*{\argmax}{arg\,max}
\usepackage{cite}
\usepackage[]{hyperref}

\title{\LARGE \bf
Active Learning based on Data Uncertainty and Model Sensitivity}

\author{Nutan Chen, Alexej Klushyn, Alexandros Paraschos, Djalel Benbouzid, Patrick van der Smagt\\
  AI Research,
  Data:Lab,
  Volkswagen Group,
  Munich, Germany \\
  \texttt{ \{first name dot last name\}@volkswagen.de} \\
}

\begin{document}

\maketitle

\input{sections/abstract}
\input{sections/introduction}

\input{sections/related_work}
\input{sections/methods}

\input{sections/experiments}

\input{sections/conclusion}

\section*{Acknowledgment}
We are very grateful to Justin Bayer for valuable suggestions concerning this work.

\bibliographystyle{IEEEtran}
\bibliography{active_learning.bib}

\input{sections/appendix}

\end{document}

%% file: sections/abstract.tex
\begin{abstract}

Robots can rapidly acquire new skills from demonstrations.
However, during generalisation of skills or transitioning across fundamentally different skills, it is unclear whether the robot has the necessary knowledge to perform the task.
Failing to detect missing information often leads to abrupt movements or to collisions with the environment.
Active learning can quantify the uncertainty of performing the task and, in general, locate regions of missing information. 
We introduce a novel algorithm for active learning and demonstrate its utility for generating smooth trajectories. 
Our approach is based on deep generative models and metric learning in latent spaces. 
It relies on the Jacobian of the likelihood to detect non-smooth transitions in the latent space, i.e., transitions that lead to abrupt changes in the movement of the robot.
When non-smooth transitions are detected, our algorithm asks for an additional demonstration from that specific region. 
The newly acquired knowledge modifies the data manifold and allows for learning a latent representation for generating smooth movements.
We demonstrate the efficacy of our approach on generalising elementary skills, transitioning across different skills, and implicitly avoiding collisions with the environment.
For our experiments, we use a simulated pendulum where we observe its motion from images and a 7-DoF anthropomorphic arm.

\end{abstract}

%% file: sections/introduction.tex

\section{Introduction}

Learning is rarely random and it typically follows an intended, often greedy curriculum.  
\emph{Actively} seeking to fill knowledge gaps is, in fact, tantamount to faster learning.  
This setup is especially advantageous when the data is scarce and expensive to acquire. 
To actively seek the missing knowledge, the learning algorithm is endowed with the ability to query an \emph{oracle} for the next datum, or the next datum to label. 
The oracle is commonly a human labeller or a demonstrator.
To this end, active learning often boils down to two components: 
a model that quantifies the learner's uncertainty and a \emph{utility} function.
The supremum of the utility function determines the next datum to be queried.
Examples of utility functions include the Information Gain~\cite{houlsby2011bayesian}, Upper Confidence Bounds~\cite{ganti2013building}, or the Expected Improvement~\cite{brochu2010tutorial}.
Active-learning approaches are efficient as, in general, require fewer data to achieve high performance~\cite{settles2010active}.
Therefore, in robotics, where data acquisition is considered expensive, utilising active learning is crucial. 
In this work, we introduce an active-learning algorithm that is primarily aimed for motion planning. 
Our approach uses latent-variable models and proposes a novel utility function that operates in the latent space.
It leverages recent advances in metric learning for latent-variable models~\cite{chen2018_aistats,arvanitidis2017latentICLR} and augments them with uncertainty estimation.

\begin{figure}[t]
    \vspace*{0.8em}
    \centering
    \begin{subfigure}[b]{0.49\columnwidth}
        \includegraphics[width=\textwidth]{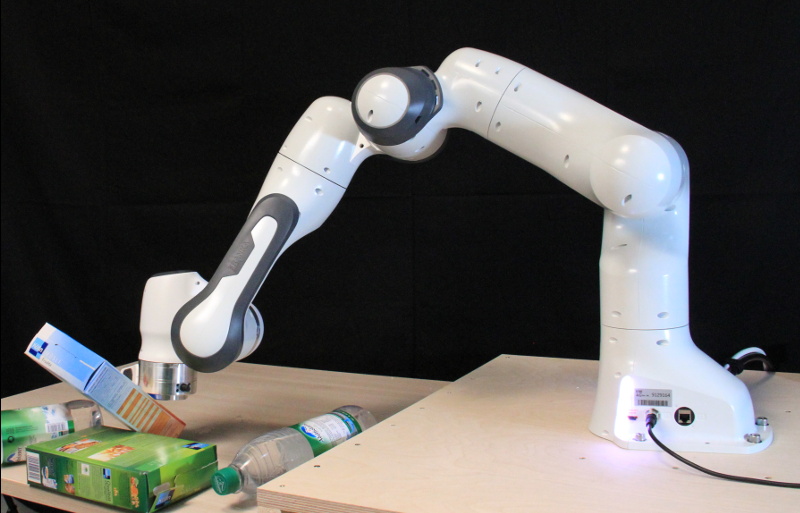}
        \caption{Before active learning.}%
    \label{fig:franka_setup_fail}
    \end{subfigure}
    \begin{subfigure}[b]{0.49\columnwidth}
        \includegraphics[width=\textwidth]{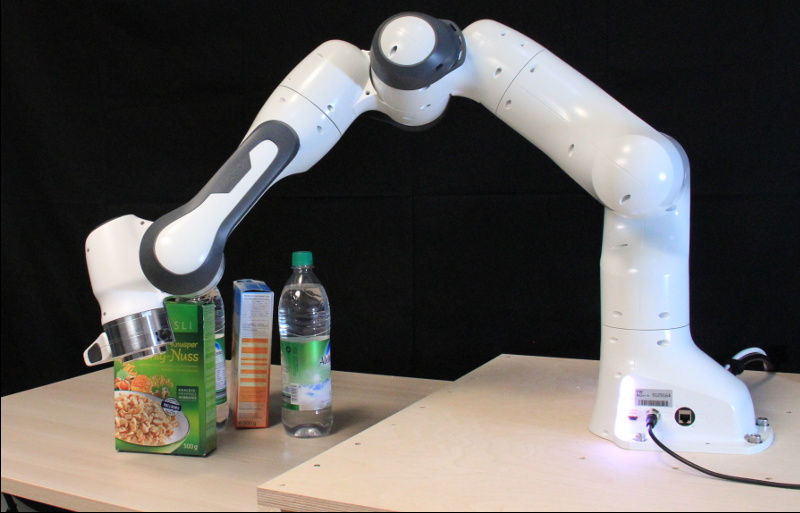}
        \caption{After active learning.}%
    \label{fig:franka_setup_suc}
    \end{subfigure}
    \caption{The FRANKA 7-DoF robotic arm performing a reaching movement.
        (\subref{fig:franka_setup_fail}) Despite that the robot was uncertain
        about performing the movement, we executed it, and, as a result, the
        robot collided with the obstacles.
        (\subref{fig:franka_setup_suc}) Our approach successfully detected that
        there is not enough information for executing the movement and asked for
        additional demonstrations. After acquiring the demonstrations,
        our approach was confident that no additional knowledge is needed and
        the robot successfully performed the reaching movement by avoiding the
        obstacles.\label{fig:firstpage}}
\end{figure}

Current approaches in active learning for robotics, further discussed in Section~\ref{sec:related_work}, mainly focus on acquiring demonstrations for learning new trajectories when data points are missing i.e., they can successfully quantify the missing information.
However, they do not provide a data-driven interpolation approach between the already acquired data points, e.g., when generalising skills or transitioning between skills.
In practice, a shortest path approach is used, but it suffers from limitations. 
First, the interpolation does not follow the shortest path in the data manifold and consequently, might drive the robot away from the known state-space regions that have been acquired from demonstrations, resulting, e.g., in collisions or in reaching joint limits.
Second, the smoothness of the movement is also not taken into account.
Yet, smooth and predicable motion is a key ingredient for safe interactions with robots.

We propose a novel active-learning algorithm that supports data-driven generalisation by allowing interpolations in the data manifold, while it is capable to simultaneously detect non-smooth, abrupt changes. 
Additionally, it quantifies the uncertainty during the interpolation and, hence, suggests new demonstrations to be provided by an oracle.
Specifically, our approach leverages from latent-variable models in order to infer a meaningful representations of the motion trajectories and exploits the Jacobian of the data likelihood along the movement to discover abrupt motions.
We demonstrate the benefits of our approach in a set of experiments by generating smooth generalisations of movements and, in addition, we demonstrate how the uncertainty of a movement can be used to implicitly avoid obstacles.
For the experimental evaluation, we model the motion of a pendulum assuming that we have access only to image observations and we use a 7-DoF anthropomorphic arm to demonstrate our approach on avoiding obstacles.

%% file: sections/related_work.tex
\section{Related work} 
\label{sec:related_work}

Active learning is a common component in robotic systems, especially when it comes to efficiently acquiring new samples during the learning 
\cite{atlas1990training,cohn1996active}, or to performing exploration in the action space \cite{salganicoff1996active, morales2004active}.
Interested by the problem of teaching robots by humans, the authors in \cite{chao2010transparent} leverage the uncertainty of the hypothesis space in order to efficiently request demonstrations from a human operator.
In \cite{cakmak2012designing}, the robot is endowed with the ability to ask questions, in order to acquire new labels, new demonstrations, or new skill representations.
The uncertainty estimation of Gaussian Processes is used in \cite{maeda2017active} in order to learn to broaden the robot reaching skills by querying new demonstrations whenever the uncertainty reaches a specified threshold.
Active learning is also used to improve over random exploration for grasping tasks based on visual sensory input \cite{salganicoff1996active}. 
Also for grasping, active learning is combined with reactive control in order to explore interesting poses using an upper confidence bound (UCB) policy \cite{kroemer2010combining}.
In \cite{baranes2013active}, a goal-driven active learning approach is developed for learning skills in continuous sensorimotor spaces.
In \cite{HanglDBP17}, the authors combine model-free and model-based reinforcement learning methods---and the uncertainty thereof, in order to acquire robotic manipulation skills.

Our robot experiments involves a query-based learning system. The scenario is similar to \cite{maeda2017active}, however, in the latter work the user has to manually choose the trigger.

More generally, active learning has been intensively studied in the machine learning literature \cite{settles2010active}. In the latter survey, the author distinguishes three types of scenario, depending on how to select the query to be labelled by the oracle; namely \emph{membership queries synthesis}, \emph{stream-based selective sampling}, and \emph{pool-based sampling}. 
In the membership queries synthesis scenario (e.g., \cite{angluin1988queries, king2009automation}), the learner first generates (or synthesises) the query to be annotated, instead of sampling it from an observed pool of data. 
In stream-based selective sampling (e.g., \cite{atlas1990training}), the learner further \emph{filters} the generated queries to be labelled and hence can decide to discard it based on a given ``informativeness measure''.
Finally, pool-based sampling (e.g., \cite{lewis1994sequential, Gal2017Active}) is motivated by applications wherein a large amount of unlabelled data can be collected but labelling by the oracle is costly. 
In our illustrative example in Section~\ref{toy_experiment}---the pendulum experiment---pool-based sampling is used. 
The original problem of the experiment corresponds to an unsupervised learning task and does not require labels, however the setup allows us to select the most useful data from the pool and evaluate our method. 
In Section~\ref{pendulum_experiment} and \ref{robot_experiment} however, stream-based selective sampling is used. These experiments imply robot interaction and it remains expensive to obtain unlabelled data for such manipulations.

In order to model the uncertainty, kernel-based methods are commonly used in the active learning literature, such as Support Vector Machines and ``margin-based uncertainty'' \cite{joshi2009multi}---when dealing with low-dimensional data.
In \cite{li2013adaptive}, the authors combine Radial Basis Functions (RBF) kernels with information density. 
Bayesian Neural Networks are also used for active learning, as recently shown by \cite{Gal2017Active} in the context of images classification. 
Following \cite{pmlr_v51_que16}, we use an RBF network to model data uncertainty.
Our method is an alternative for modelling defect detection by measuring model sensitivity. 

%% file: sections/methods.tex
\section{Applying Riemannian geometry to latent variable models for active learning}
Latent-variable models (LVM), defined by
\begin{equation}
\label{eq:nlvm}
\p{\x} = \int \p{\x}{\z}\,\p{\z}\, d\z,
\end{equation}
are widely used to find a representation of observable data $\x \in \mathbb{R}^{N_x}$ through latent variables $\z \in \mathbb{R}^{N_z}$ based on hidden, nonlinear regularities in $\x$. 

We use latent variables for generating a sequence of observable data points, with the condition that every generated point has a high similarity to the previous one.
However, the similarity between the successive data points depends on the information provided to the LVM.
In case of a low similarity we apply active learning to get targeted the missing information.

Gauging the similarity of two data points in the latent space is one of the main topics in this paper.
To solve this problem, we take the Jacobian of the likelihood into account by treating the latent space as a Riemannian manifold. 
The Riemannian metric defines a relationship based on the Jacobian of the likelihood, due to change of variables when moving from Riemannian (latent) to Euclidean (observation) space. 

A \emph{smooth} interpolation through our observable data can be obtained by following the geodesic, i.e.\ the length-minimising curve between two points in the Riemannian space. 
Here, \emph{smooth} refers to a strong similarity of successive data points.

However, even when following the geodesic, finding a smooth interpolation will fail under certain circumstances. 
For instance, when we are trying to interpolate between different classes.
The distance between the data manifolds of the different classes in the observation space typically results in a high Jacobian value of the likelihood mean when interpolating from one class to the other.
This implies that information is missing to provide a sequence of similar data points to connect the different manifolds smoothly. 
As a consequence, the variance of the likelihood changes as well.

Since the Jacobians of both the mean and the variance are taken into account by the Riemannian metric, this property can be turned to advantage when dealing with active learning.
For instance, when trying to interpolate between different robot movements. 
Because missing data can be queried specifically if such boundaries are passed.

Building on that, the focus of our paper lies on applying Riemannian geometry to LVMs for active learning of robot movements.

\subsection{Importance-weighted autoencoder}
\label{sec:methods-iwae}
Since in most LVMs the integral in Eq.~(\ref{eq:nlvm}) is intractable, approximations are used which base on sampling \cite{hastings1970monte}\cite{gelfand1990sampling} or on variational inference \cite{KingmaW13}\cite{rezende2014stochastic}.
In the latter case, the problem is reformulated as the maximisation of the evidence lower bound (ELBO). 
The distribution $\q{\z}$ approximates the intractable posterior and $p_\theta(\x|\z)$, defined as the generative model and parameterised by $\theta$, approximates the likelihood. 
Let $\mathbf{X} = \{ \x^{(1)},\dots,\x^{(N)} \}$ be observable data and $\z^{(i)}$ the corresponding latent variables. 
Then,
\begin{equation}
\begin{aligned}
\ln p_{\theta}(\mathbf{X})
\geq \sum_{i=1}^{N} \mathbb{E}_{q(\z^{(i)})} \Big{[} \ln  \frac{p_{\theta}(\x^{(i)} | \z^{(i)}) \, p_{\theta}(\z^{(i)})}{q(\z^{(i)})} \Big{]} = \elbo .
\end{aligned}
\end{equation}
Implementing $\q{\z^{(i)}} = q_{\phi}(\z^{(i)}|\x^{(i)})$ with a neural network parameterised by $\phi$, we obtain the variational autoencoder (VAE) introduced in \cite{KingmaW13, rezende2014stochastic}.

To overcome the limitations of ordinary VAEs and to achieve a tighter ELBO, we use importance-weighted autoencoders (IWAE) \cite{BurdaGS15, cremer2017reinterpreting} in our approach. 
IWAEs treat $q_{\phi}(\z | \x)$ as a proposal distribution and obtain a tighter ELBO by using importance sampling:
\begin{equation}
\label{eq:K}
\begin{aligned}
& \elbo = \sum_{i=1}^{N} \mathbb{E}_{\z^{(i)}_{1},\dots , \z^{(i)}_{K} \sim q_{\phi}(\z^{(i)} | \x^{(i)})} \Big{[} 
\ln \frac{1}{K} \sum_{k=1}^{K} w^{(i)}_{k} \Big{]},
\end{aligned}
\end{equation}
with the importance weights
\begin{equation}
w^{(i)}_{k} = \frac{p_{\theta}(\x^{(i)} | \z^{(i)}_{k}) \,p_{\theta}(\z^{(i)}_{k})}{q_{\phi}(\z^{(i)}_{k} | \x^{(i)})}.
\end{equation}

\subsection{Riemannian geometry in latent variable models}
\label{sec:methods-riemannian-geometry}
Riemannian space is a differentiable manifold $M$ which contains as an additional characteristic a metric to describe its geometric properties. 
The corresponding metric tensor $\G$ assigns to each point $\mathbf{\z}$ in the latent space an inner product on the tangent space $T_{\mathbf{\z}} M$, defined by
\begin{equation}
\langle\z', \z'\rangle_{\z} := \z'^{T}\, \G(\z)\, \z',
\end{equation}
with $\z' \in T_{\z}M$ and $\z \in M$.

Let us assume we have a curve $\cz:[0, 1]\rightarrow\mathbb{R}^{N_{z}}$ in the Riemannian (latent) space that is transformed by a continuous function $\cx(\cz(t))$ to an $N_{x}$-dimensional Euclidean (observation) space, where $\cz(t)\in\mathbb{R}^{N_{z}}$.
The length of this curve in the Euclidean space is defined as
\begin{align} \label{eq:length_final}
L(\cz) = \int_0^1\sqrt{\big<\dot{\gamma}(t), \dot{\gamma}(t)\big>_{\gamma(t)}}\mathrm{d}t,
\end{align}
with the metric tensor $\G=\mathbf{J}^{T}\mathbf{J}$, where $\mathbf{J}$ is the Jacobian matrix of the likelihood and $\dot{\gamma}$ the time derivative of $\gamma$.

\subsection{Using geodesics for trajectory generation}
\label{sec:methods-geodesic}
To approximate the geodesic we use a neural network that is optimised by minimising $L(\cz)$. A singular-value decomposition of $\G$ ensures the geodesic is following the data manifold, as introduced in \cite{chen2018_aistats}.

Although this method takes the sensitivity of the model into account, it does not capture data uncertainty. 
In other words: 
the high Jacobian values of the likelihood mean at the boundaries between different data manifolds are taken into account, but there is nothing that tells us where our model is uncertain due to missing data.
The reason is a global variance of the generative model. 
To remedy that, the neural network of the generative model is extended by radial basis function (RBF) networks to be able to represent the likelihood variance too \cite{arvanitidis2017latentICLR, pmlr_v51_que16}.

In contrast to \cite{arvanitidis2017latentICLR}, we update the weights of the RBF networks during the training of the generative model and define a rule for an autonomous hyperparameter selection after the training is finished. 
The RBFs $\mathbf{v}$ and the precision $\bm{\psi}(\z)$ of the generative model are given by
\begin{align}
\bm{\psi}(\z) = \mathbf{W}\mathbf{v}(\z), \\
v_k(\z) = \exp \bigl(-\lambda_k \| \z - \mathbf{c}_k \| _2\bigr), \ & \text{with } k = 1, \dots , K \nonumber,
\end{align}
where $K$ is the number of the radial basis functions. $\lambda$ and $\mathbf{c}$ are variables representing bandwidth and centres, respectively. 
$\mathbf{W}$ are the weights to be optimised. 
The bandwidth is defined by
\begin{align}
\lambda_k = \alpha \left(\frac{\sum^K_i \|\mathbf{c}_k -\mathbf{c}_i \|_2}{k}\right)^{-2},
\end{align}
where $\alpha$, a hyperparameter, denotes the curvature of the Riemannian metric. 
Since the variance is the reciprocal of the precision:
\begin{align}
\bm{\sigma}^2(\z) = \bm{\psi}(\z)^{-1},
\end{align}
it increases with the distance to the centres and the uncertainty of the model, respectively. 
It is not possible to directly compute the Jacobian of a sample $\x\sim p_\theta(\bobs|\blatent)$. 
Hence, we reparameterise it by $\epsilon\sim\mathcal{N}(0, 1)$ \cite{KingmaW13, RezendeMW14}:
\begin{align}
\J(\z) = \J_\mu(\z)+\epsilon~\J_\sigma(\z),
\end{align}
where $\J_\mu$ and $\J_\sigma$, the Jacobians of the mean and the standard deviation of the likelihood, represent the sensitivity and the data uncertainty of the model, respectively. 
The changes in the likelihood variance have influence on $\G$, hence the equation introduced in \cite{chen2018_aistats} has to be updated. 
To simplify the calculation, we remove the stochasticity in $\G$ by taking the expectation \cite{arvanitidis2017latentICLR}
\begin{align}
\mathbb{E}_{p(\epsilon)}[\G(\z)] = \J_{\mu}(\z)^T \J_\mu(\z) + \J_{\sigma}(\z)^T\J_{\sigma}(\z).
\end{align}

We differ from \cite{arvanitidis2017latentICLR} in the optimisation procedure of the model: 
the centres $\mathbf{c}$ are computed by K-means and updated at every $n$-th iteration step during the training of the IWAE.
For both the computation of the centres $\mathbf{c}$ and the RBFs $\mathbf{v}$, the mean $\z_\mu$ of $\z$ is used.
The weights $\mathbf{W}$ are optimised by back-propagation. 
$\alpha$ is treated as a hyperparameter during the IWAE training. 
After the training is finished, $\alpha$ is updated to satisfy
\begin{align}
\| \max [ \J_\mu (\z_\mu) ] - \max [ \J_\sigma (\z_\mu) ] \| < \epsilon, \text{ with } \epsilon\rightarrow 0.
\label{eq:upal}
\end{align}
Satisfying Eq.~(\ref{eq:upal}) guarantees that the mean and the variance have a similar effect on the Riemannian metric tensor.

\begin{figure*}[t]
    \centering
    \newsavebox{\largestimage}
    \savebox{\largestimage}{%
        \includegraphics[width=0.65\columnwidth]{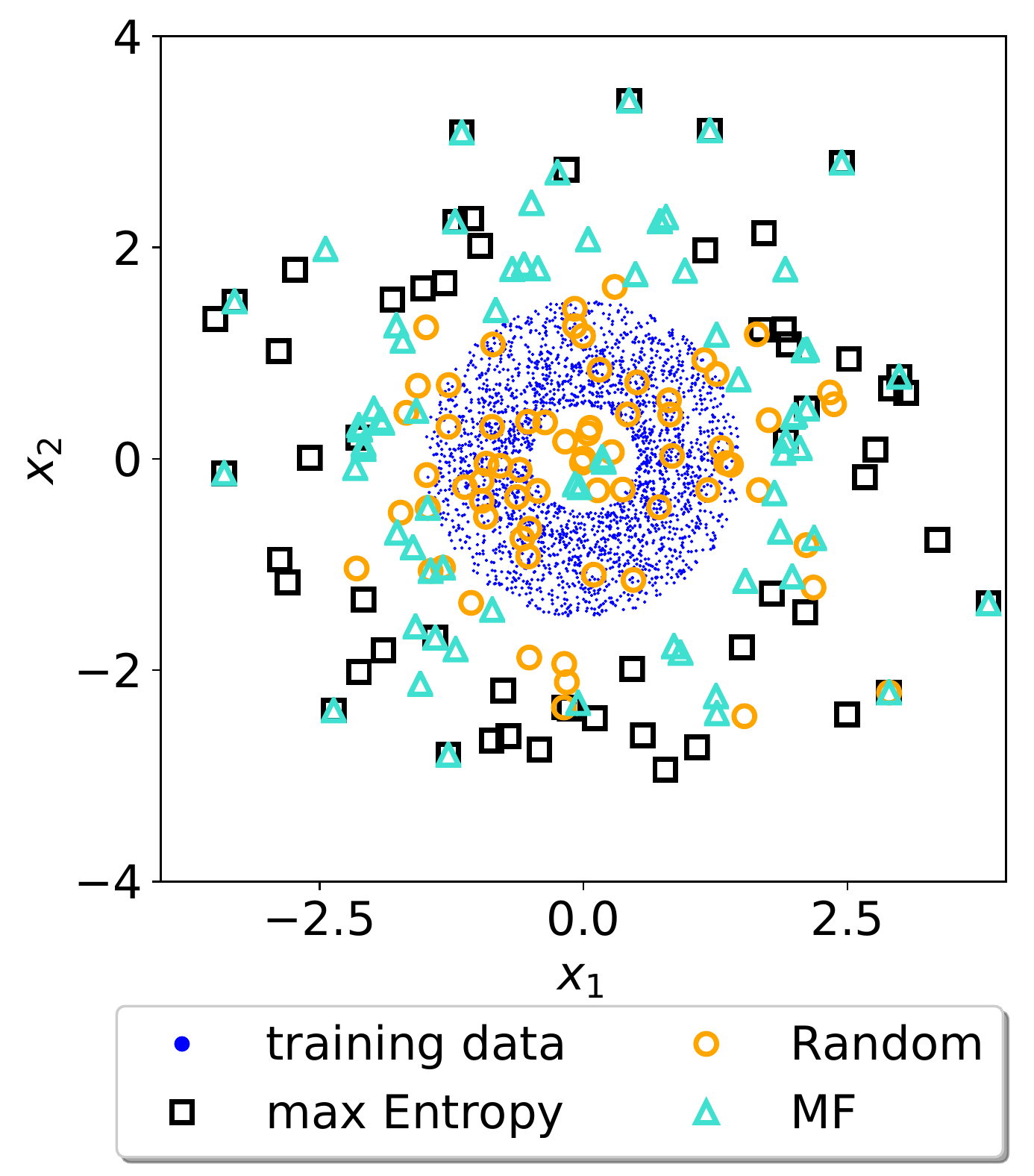}
    }%
    \begin{subfigure}[b]{0.65\columnwidth}
        \centering
        \raisebox{\dimexpr.5\ht\largestimage-.45\height}{%
            \includegraphics[width=\columnwidth]{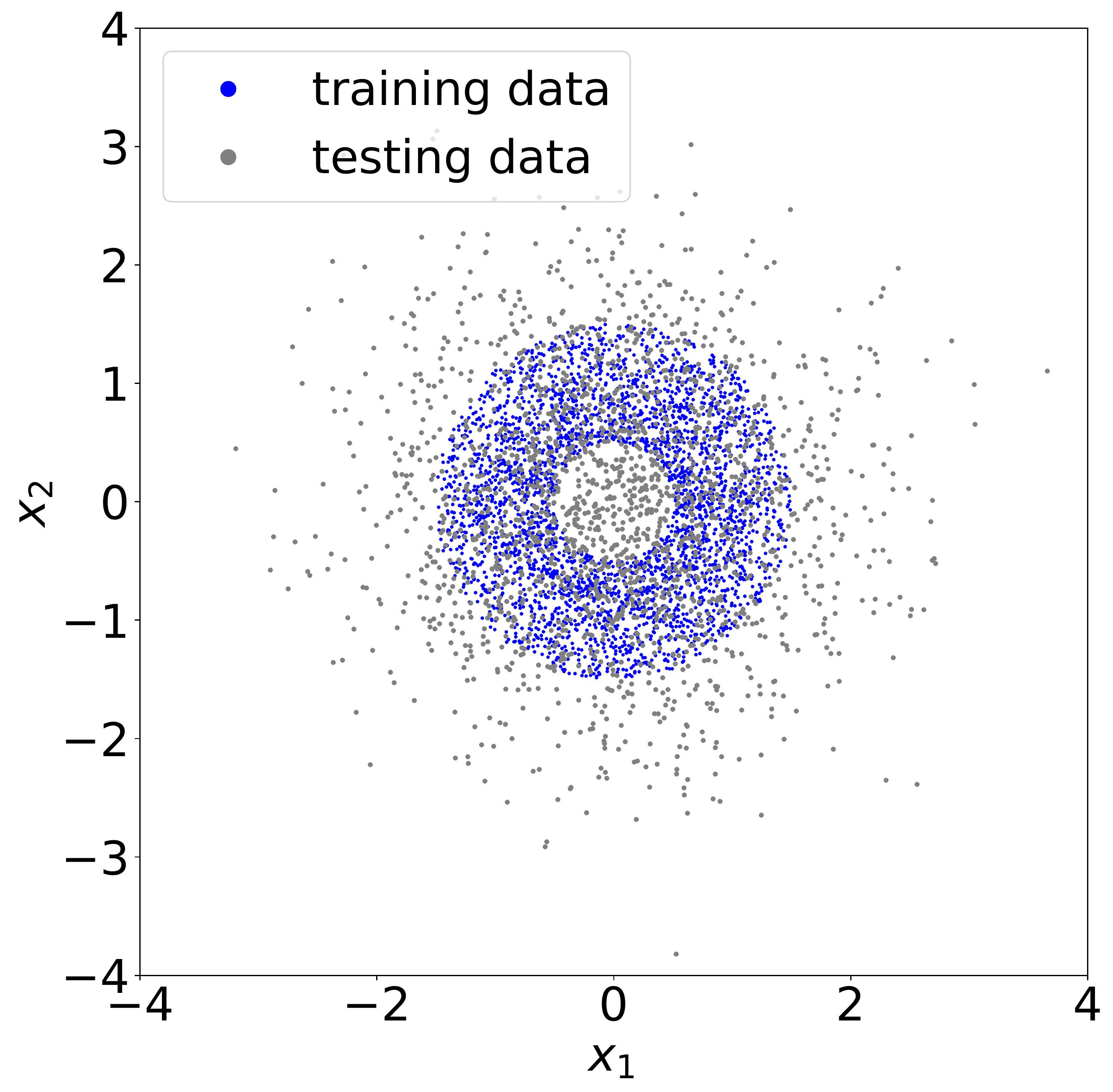}
        }
        \caption{Training and testing datasets}%
        \label{fig:toy_data_a}
    \end{subfigure}%
    \hfill%
    \begin{subfigure}[b]{0.65\columnwidth}
        \centering
        \raisebox{\dimexpr.5\ht\largestimage-.5\height}{%
            \includegraphics[width=\columnwidth]{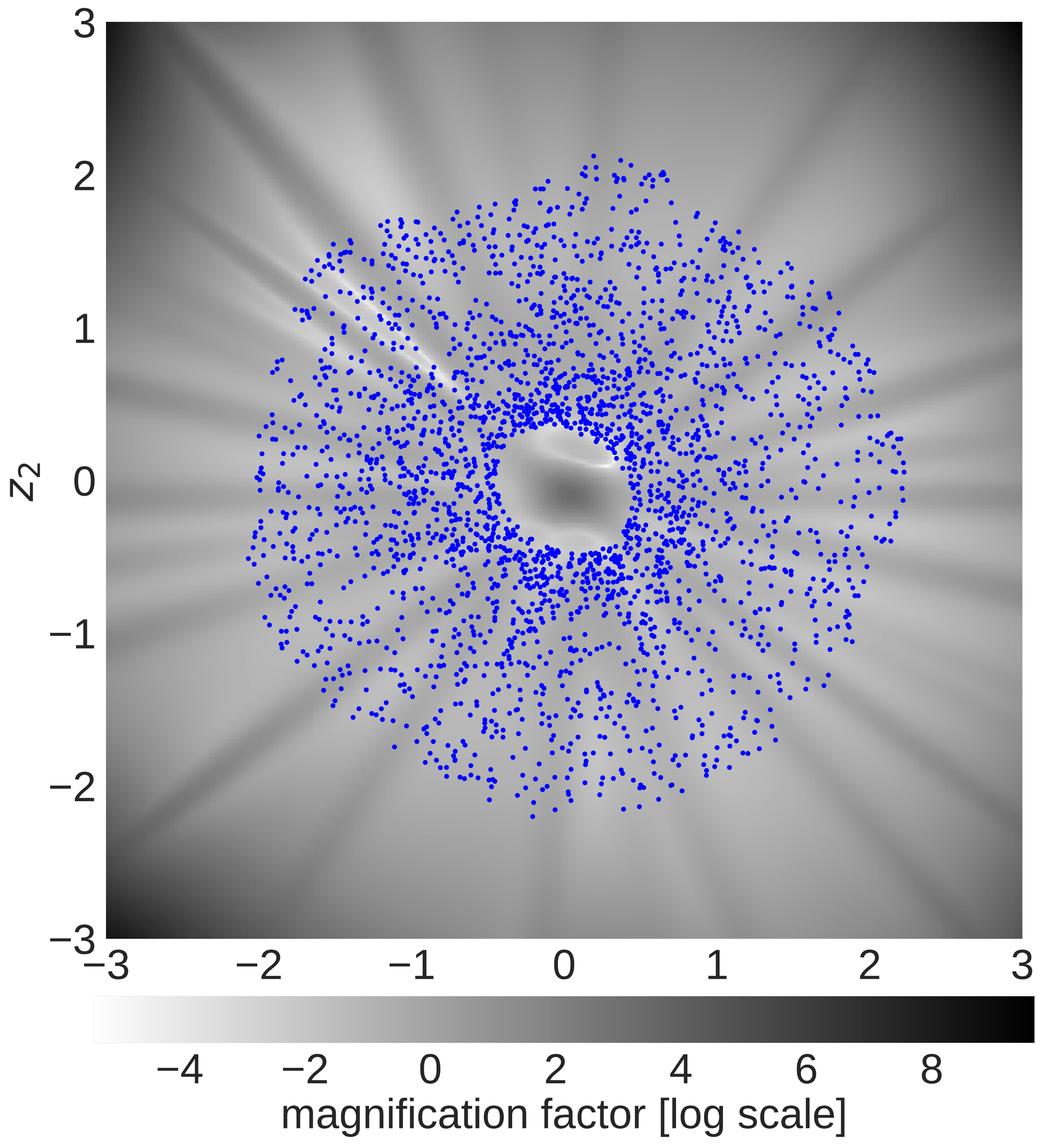}
        }
        \caption{$\MF$ and training data in the latent space}%
        \label{fig:toy_data_b}
    \end{subfigure}%
    \hfill%
    \begin{subfigure}[b]{0.65\columnwidth}
        \centering
        \raisebox{\dimexpr.5\ht\largestimage-.5\height}{%
            \includegraphics[width=\columnwidth]{toy_methods_acquired}
        }
        \caption{Acquired data points}%
        \label{fig:toy_data_c}
    \end{subfigure}%
    \\
    \caption{Evaluations of the illustrative experiment based on a two-dimensional dataset.
    (\subref{fig:toy_data_a})
    Training and testing dataset with a sample sizes of $3.3\cdot10^3$ and $2\cdot10^3$ data points, respectively.
    The data-acquisition pool has the same distribution as the testing dataset and $2\cdot10^3$ data points.
    (\subref{fig:toy_data_b})
    Latent space of the trained model. The $\MF$ is represented by the grayscale. 
    The blue points depict the mean of the training data.
    (\subref{fig:toy_data_c})
    Observation space of the acquired data after seven iterations when using different active-learning approaches, namely the $\MF$, the Max Entropy, and a random acquisition strategy. 
    The latter acquires data that is similarly distributed to the data in the acquisition pool, which has a large overlap with the training dataset and, therefore it does not provide an efficient learning approach.  
    The Max Entropy does not take into account the $\MF$ of the model or acquires data points from the center of the latent space.
    }%
    \label{fig:toy_data}
\end{figure*}

\subsection{Active learning for robot trajectory generation}
\label{sec:methods-active-learning}
Active learning can be applied to targeted reduce the uncertainty of our model, which leads to smoother trajectory generations.
In active learning an acquisition function $a$ is used to detect where a model $M$ is uncertain, so missing labels can be queried specifically:
\begin{align}
\x^{\ast} = \argmax_{\x\in D_\mathrm{pool}} a(\x, M).
\label{eq:al}
\end{align}
Our goal is to guarantee a smooth interpolation along the geodesic. 
This is only possible if there are no abrupt changes in the Jacobian of the likelihood, which is expressed by the determinant of the metric tensor $\G$ for a specific pair $(\x, \z)$.
This leads to the following acquisition function:
\begin{align}
a(\x, M) = \sqrt{\det\G(\z)} \eqqcolon \MF(\z),
\label{eq:acqfct}
\end{align}
also defined as the magnification factor ($\MF$) \cite{bishop1997magnification}.
The $\MF$ can be interpreted as the scaling factor when moving from the Riemannian (latent) to the Euclidean (observation) space, due to the change of variables.

In addition to the acquisition function, a threshold is necessary to tell the active learning algorithm whether a interpolation is smooth or not. 
The threshold is defined as
\begin{align}
\tau(\Omega) = \frac{1}{N_\Omega}\sum_{i=1}^{N_\Omega}\omega_{i}+\sqrt{\text{Var}(\Omega)}, 
\end{align}
where $\omega_{i}\in \Omega$, $\Omega=\{\MF(\z^{(1)}), \dots, \MF(\z^{(N)})\}$, and $N_\Omega$ is the cardinality of $\Omega$. 
\\

When applying active learning to robot movements, we use a set of pairs of start and end points in the observation space $\Pi = \{(\x_0^{(1)}, \x_1^{(1)}), \dots , (\x_0^{(N_\Pi)}, \x_1^{(N_\Pi)})\}$. 
For each pair the geodesic $\mathrm{geo}(\x_0, \x_1)$ is computed.
To decide whether the movement (interpolation) between a start and an end point is smooth, only points along the geodesic $\gamma(t)\in \mathrm{geo}(\x_0, \x_1)$ are taken into account.
Thus, in contrast to the active learning approach described in Eq.~(\ref{eq:al}), $D_\mathrm{pool}=[\gamma(0), \gamma(1)]$ refers to the latent space.
Based on whether the values of the magnification factor along the geodesic $\mathrm{geo}(\x_0, \x_1)$ exceed $\tau(\Omega)$, the active learning algorithm decides if the trajectory between $\x_0$ and $\x_1$ is required to be demonstrated. 
In case of a required demonstration, retraining the model with the new data leads to low $\MF$s along $\mathrm{geo}(\x_0, \x_1)$.

Hence, the final result of our approach is a smooth movement or rather a smooth combination of movements of the robot---realised by reconstructing the latent variables along the geodesic.

\begin{figure}[t]
    \vspace*{0.8em}
    \centering
        \includegraphics[width=0.4\textwidth]{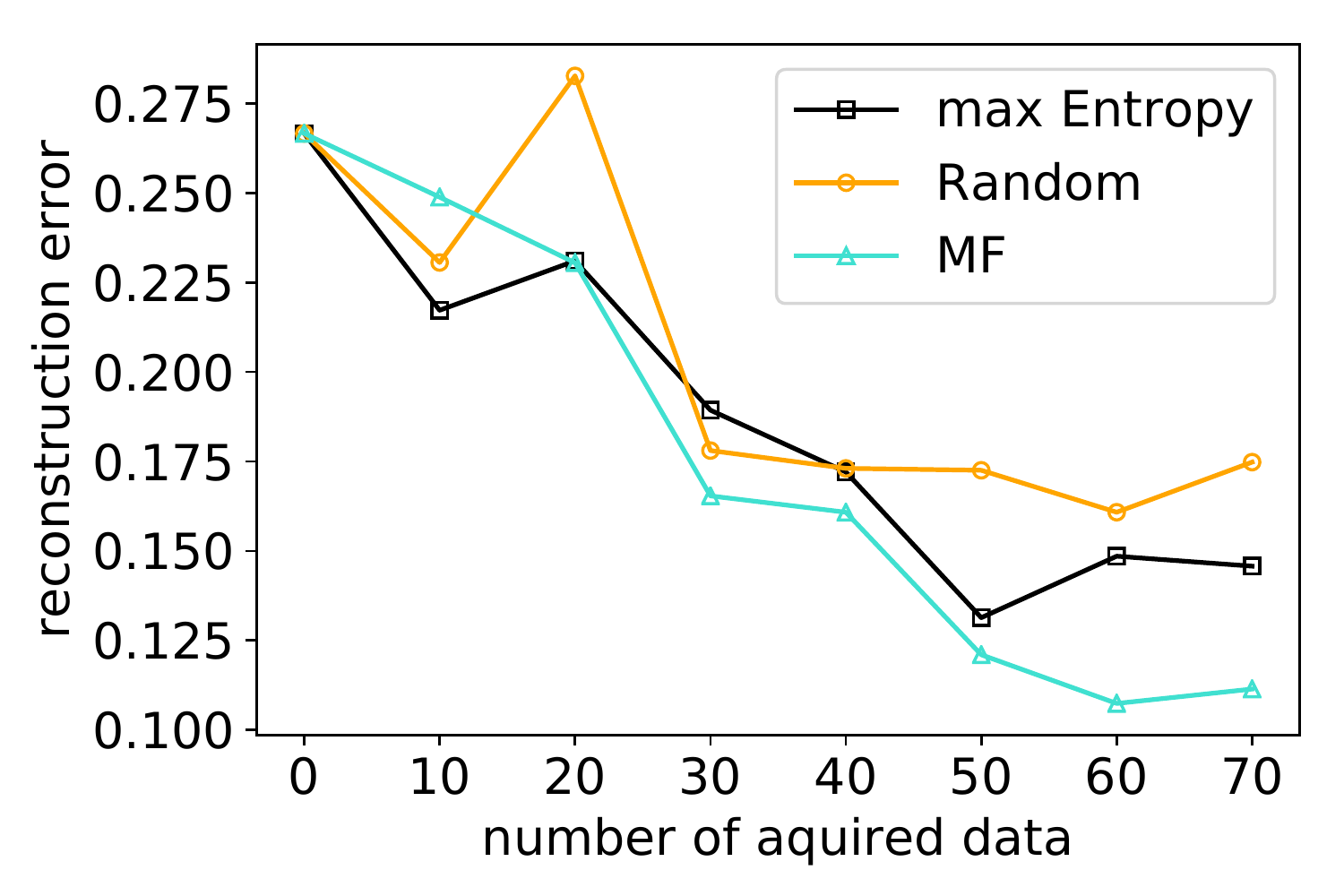}
        \caption{The reconstruction error of the illustrative experiment when using the $\MF$, Max Entropy, or random
            active-learning strategy. The acquisition functions acquire ten data points per iteration.\label{fig:toy_data_d} }
\end{figure}

%% file: sections/experiments.tex
\begin{figure*}[ht]
    \centering
    \begin{subfigure}[b]{0.3\textwidth}%
        \includegraphics[width=\textwidth]{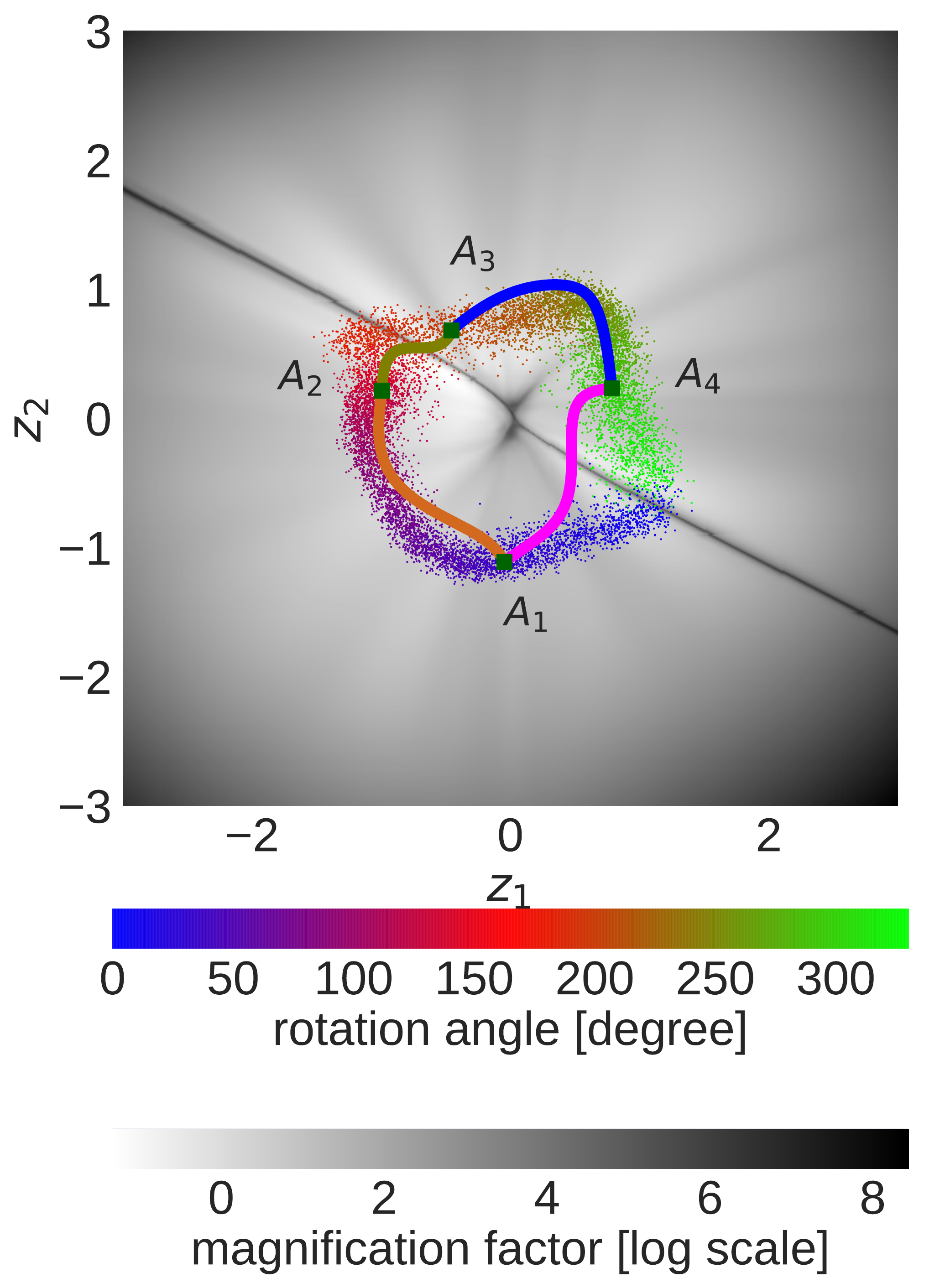}
        \caption{Latent manifold and $\MF$}%
        \label{fig:pendulum_latent_before_active}
    \end{subfigure}\hfill
    \begin{minipage}[b]{0.68\textwidth}
        \centering
        \begin{subfigure}[b]{0.75\textwidth}%
            \raisebox{\dimexpr.5\ht\largestimage-.5\height}{%
                \includegraphics[width=\textwidth]{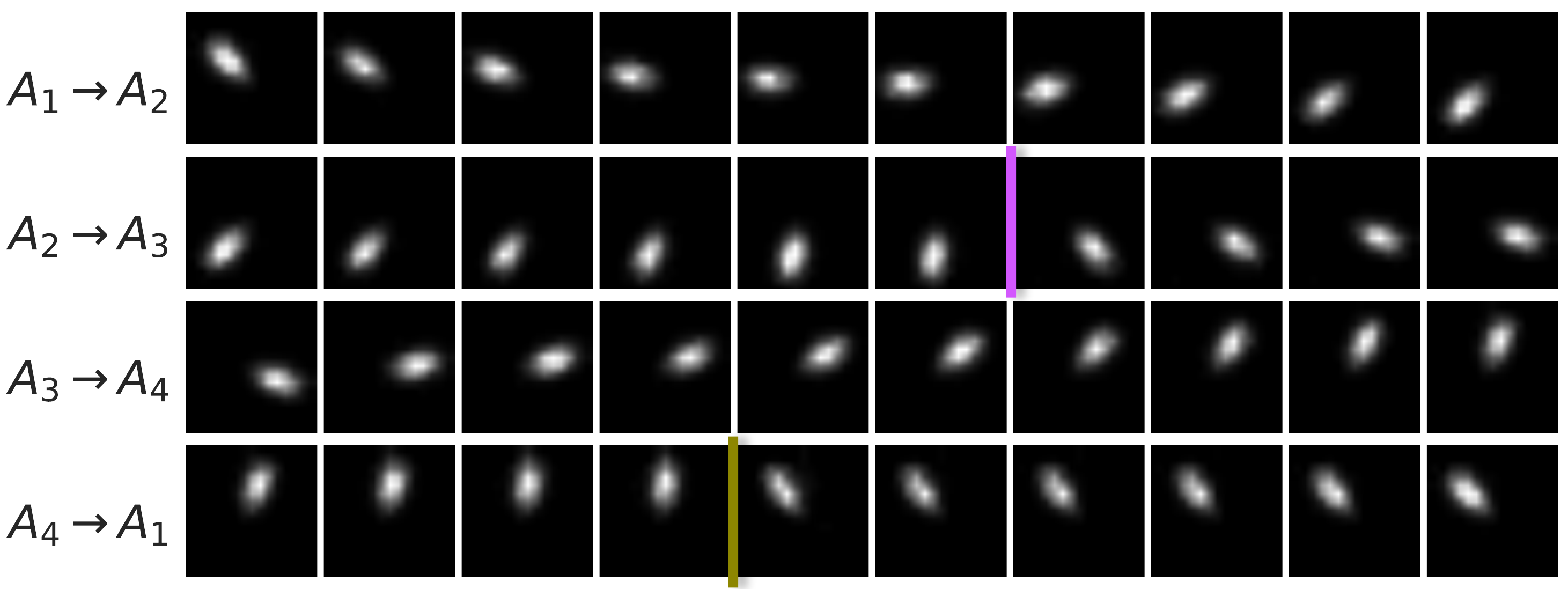}
            }
            \caption{Reconstruction of the trajectories}%
            \label{fig:pendulum_recon_before_active}
        \end{subfigure}\\[1em]
        \begin{subfigure}[b]{0.65\textwidth}%
            \hspace{10pt}
            \raisebox{\dimexpr.5\ht\largestimage-.5\height}{%
                \includegraphics[width=\textwidth]{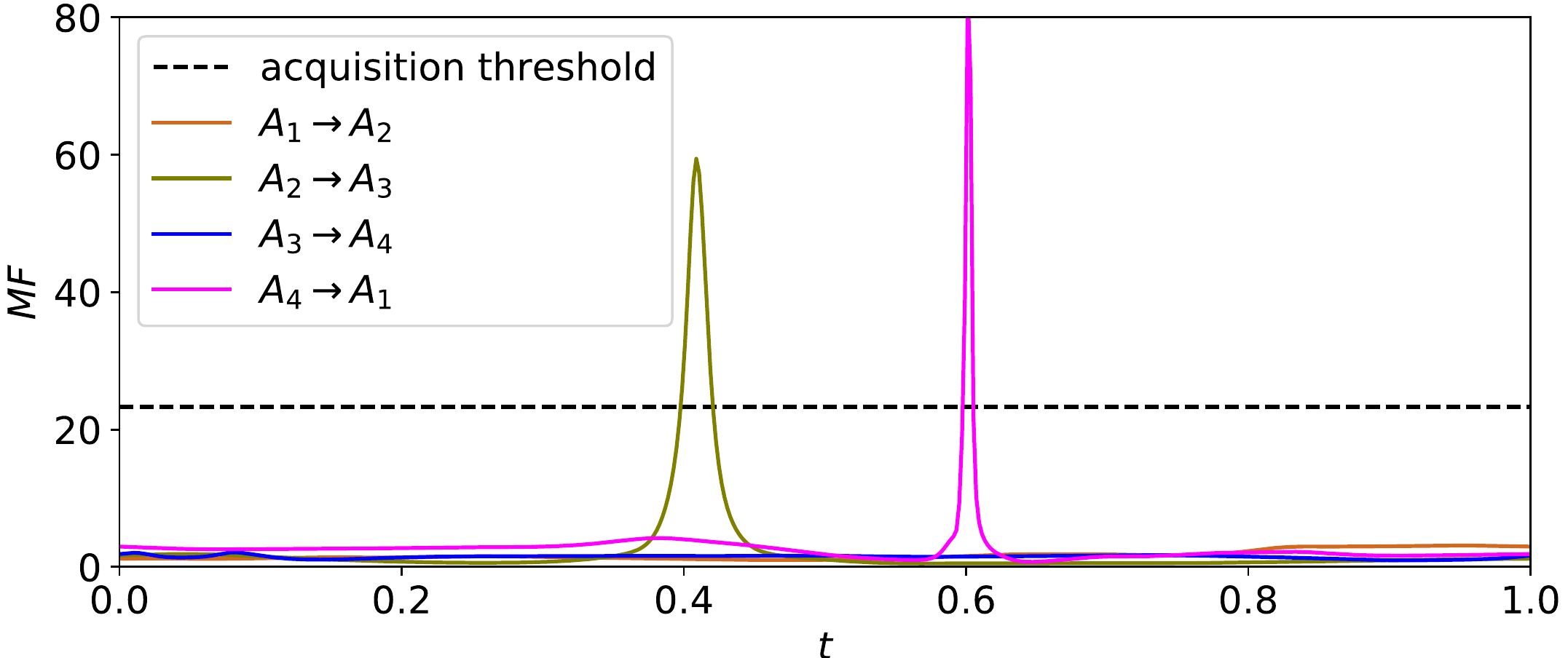}
            }
            \caption{$\MF$ during the trajectory execution}%
            \label{fig:pendulum_mf_before_active}
        \end{subfigure}
    \end{minipage}
    \caption{%
        Evaluation of the pendulum experiment \emph{before} active learning. %
        (\subref{fig:pendulum_latent_before_active})
        Latent space of the pendulum dataset after the initial training.
        The markers $\{A_1, A_2, A_3, A_4\}$ represent the position of the pendulum at $\{40, 120, 200, 280\}$ degrees, respectively. 
        In addition, the colour encodes the pendulum rotation angles and the greyscale the $\MF$ values.
        (\subref{fig:pendulum_recon_before_active})
        The trajectories $A_2\rightarrow A_3$ and $A_4\rightarrow A_1$ of the generated pendulum movements experience large $\MF$ values, leading to discontinuities. 
        The discontinuities are marked by coloured lines.
        (\subref{fig:pendulum_mf_before_active})
        The $\MF$ exceeds the threshold for two trajectories, $A_2\rightarrow A_3$ and $A_4\rightarrow A_1$.
    }%
\end{figure*}

\begin{figure*}[t]
    \centering
    \begin{subfigure}[b]{0.3\textwidth}%
        \includegraphics[width=\textwidth]{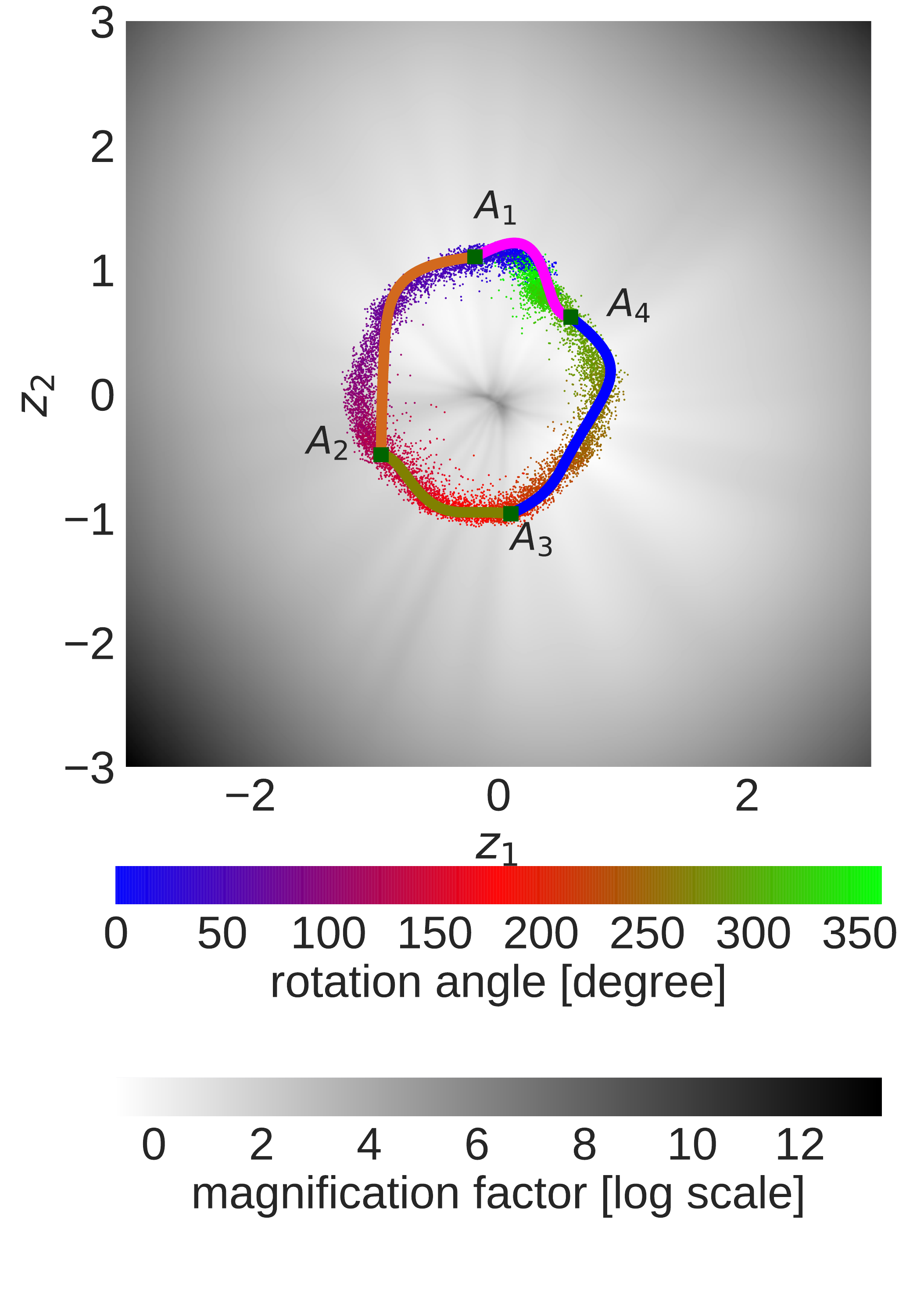}
        \caption{Latent manifold and $\MF$}%
        \label{fig:pendulum_latent_after_active}
    \end{subfigure}
    \hfill
    \begin{minipage}[b]{0.68\textwidth}
        \centering
        \begin{subfigure}[b]{0.75\textwidth}%
            \raisebox{\dimexpr.5\ht\largestimage-.5\height}{%
                \includegraphics[width=\textwidth]{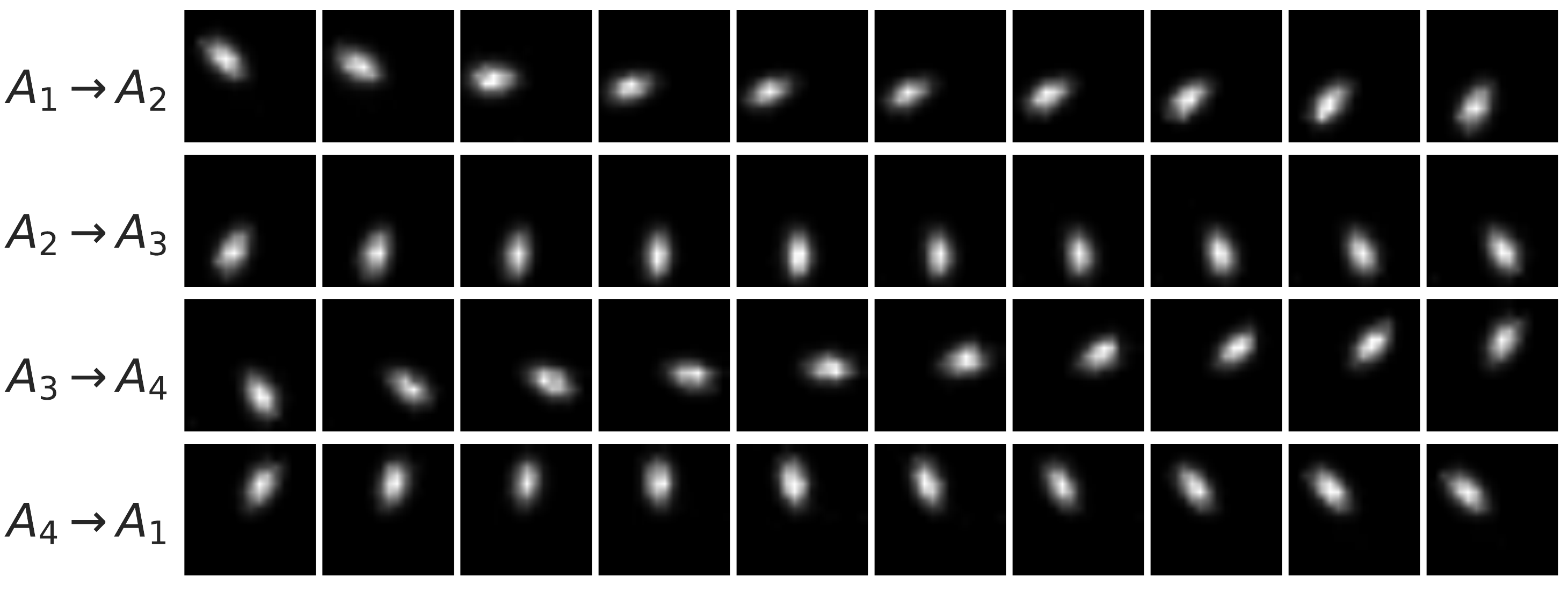}
            }
            \caption{Reconstruction of the trajectory}%
            \label{fig:pendulum_recon_after_active}
        \end{subfigure}\\[1em]
        \begin{subfigure}[b]{0.65\textwidth}%
            \hspace{10pt}
            \raisebox{\dimexpr.5\ht\largestimage-.5\height}{%
                \includegraphics[width=\textwidth]{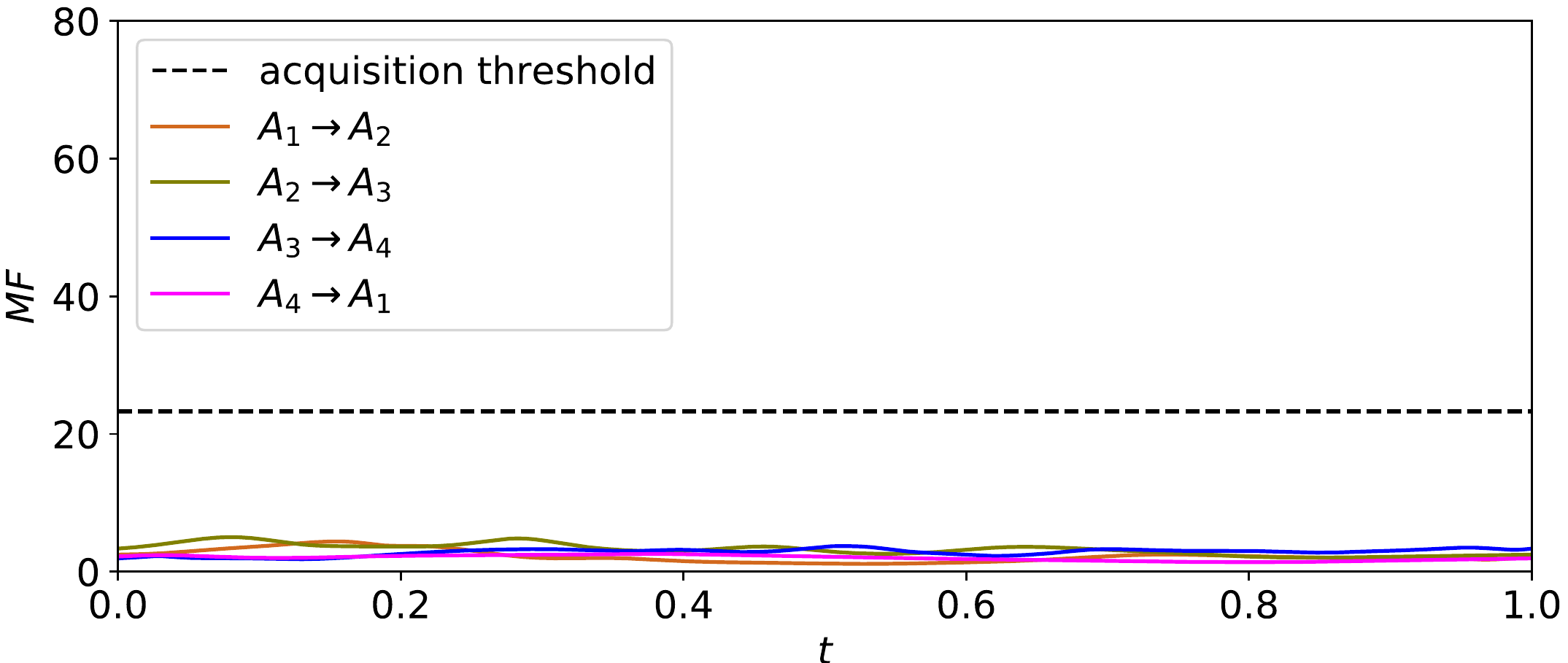}
            }
            \caption{$\MF$ during the trajectory execution}%
            \label{fig:pendulum_mf_after_active}
        \end{subfigure}
    \end{minipage}
    \caption{%
        Evaluation of the pendulum experiment \emph{after} active learning. %
        (\subref{fig:pendulum_latent_after_active})
        The resulting latent space of the pendulum dataset shows that the $\MF$ decreased along the planned trajectories.
        (\subref{fig:pendulum_recon_after_active})
        The generated pendulum movements are smoother and they do not experience high $\MF$ values.
        (\subref{fig:pendulum_mf_after_active})
        The $\MF$ values of all four generated pendulum trajectories are below the threshold.
    }%
\end{figure*}

\section{Experimental Evaluation}

We evaluate our approach in multiple scenarios. 
First, we use an artificial two-dimensional dataset to illustrate how our approach works.  
Then, we demonstrate that our approach can work efficiently with high-dimensional data on simulated pendulum where the state is given by images. 
Finally, we present our results on controlling a 7-DoF robotic arm where smooth reaching movements are generated. 
When our approach detects that a trajectory would cross regions of the state space where not enough data
have been acquired, it asks for additional demonstrations. 
Hence, it is used to implicitly avoid collisions and joint limits.
The architectural design and the hyper-parameters used for our experiments are listed in the appendix.

\subsection{Illustrative experiment}\label{toy_experiment}

In the first experiment, we evaluate the efficiency of our approach in reducing the reconstruction error by actively acquiring data points from regions where the model does not have enough information. 
To better illustrate our approach, we generated an artificial two-dimensional dataset, where the IWAE maps the observation space to a two-dimensional latent space. 
The training dataset is depicted in Fig.~\ref{fig:toy_data_a}, whereas the resulting latent space and $\MF$ are shown in Fig.~\ref{fig:toy_data_b}.

Our algorithm asks for new data points from regions where the $\MF$ is high, and therefore it selects the central region. 
In contrast, the Max Entropy strategy assumes that enough information form the central region is present.  

As a result, our $\MF$ active-learning approach can efficiently reduce the reconstruction error using fewer samples than the Max Entropy or the random acquisition approach. 
The results are shown in Fig.~\ref{fig:toy_data_d}.

\subsection{Trajectory planning for pendulum}\label{pendulum_experiment}

We demonstrate the trajectory planning capabilities of our approach in a simulated 1-DoF pendulum system.  
The simulator provides a $16 \times 16$-pixel image of the current state of the pendulum, which we use as input to our
algorithm.  
We gathered an image dataset by collecting $T=15 \cdot10^3$ images for two different joint angle ranges, $R_1 = [0, 150)$ and $R_2 = [180, 330)$ degrees.  
Subsequently, we augmented the dataset by adding $0.05$ Gaussian noise to each pixel, to avoid over fitting and to improve the coherence of the latent space.  

After training, we generated four trajectories between the two datasets by following the geodesic. 
The generated trajectories are illustrated in Fig.~\ref{fig:pendulum_latent_before_active}.  
The trajectories $A_2A_3$ and $A_4A_1$ move across regions of the state space where the $\MF$ exceeds a predetermined threshold, as not enough information has been collected from those regions. 
An illustration of the trajectories is shown in Fig.~\ref{fig:pendulum_mf_before_active}.

Our approach requested for additional demonstrations from regions where the $\MF$ exceeds the threshold.
Afterwards the model is retrained with the new data.
As a result, the $\MF$ is reduced in these regions, as shown in Fig.~\ref{fig:pendulum_latent_after_active}.  
The corresponding trajectories are significantly smoother after our active-learning approach was applied, as can be seen
in Fig.~\ref{fig:pendulum_recon_after_active}.

\subsection{Generating robot trajectories with active learning}\label{robot_experiment}

Deciding whether the robot is able to perform a task or a demonstration is required is not trivial.  
Therefore, we evaluate our approach in a robot trajectory generation setting, where the robot should consult the human
operator to avoid collisions with the environment. 
Also, the generated trajectories should not have abrupt changes to enable the robot to precisely follow them.  
For this experiment, we used a Panda robot from FRANKA, a lightweight 7-DoF robotic arm with joint torque sensors. 

For training our model, we provided demonstrations of reaching objects that were placed at two distinct locations.  
We used kinaesthetic teaching, i.e., the human demonstrator could freely move the robot to acquire a dataset of
five demonstrations per reaching location. 
The setup is depicted in Fig.~\ref{fig:firstpage}.
During the demonstrations we recorded the joint angles of the robot at a rate of $1\,\mathrm{kHz}$.  
Additionally, obstacles where placed in the workspace of the robot. 
Naively generating movements based on the demonstration dataset likely results in collisions.

We generated trajectories by computing an interpolation between the two distinct locations by following the geodesic, as shown in Fig.~\ref{fig:franka_latent_before}.
The geodesic trajectory crosses a region with high $\MF$ values.

\begin{figure*}[t]
    \centering
    \includegraphics[width=0.98\textwidth]{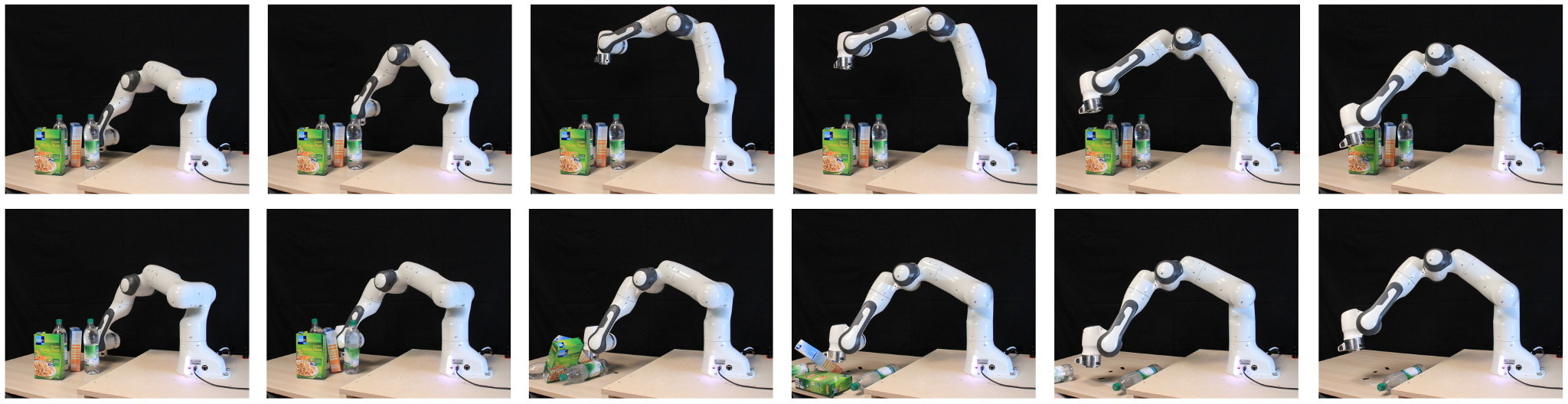}
    \caption{%
        The FRANKA robot performing a novel reaching movement. 
        (top) Our algorithm detected that an additional demonstration is required due to the lack of data. 
        After preforming active learning the robot successfully avoids the obstacle. 
        (bottom) Without the active-learning approach the robot collides with the environment.\label{fig:franka_traj_img}}
\end{figure*}

Since the $\MF$ values along the proposed trajectory exceed the threshold, the algorithm asks the user for additional data.  
Thus, after collecting the data of the queried trajectory, we retrained our model on the new data and recomputed the geodesic.
The updated latent space is shown in Fig.~\ref{fig:franka_latent_after}, where the geodesic does not cross high-$\MF$ regions anymore.
The end-effector trajectories before and after retraining are depicted in Fig.~\ref{fig:franka_traj}. 
As a result, the robot arm moves close to the demonstrated path and avoids collisions with obstacles. 
A visualisation of the robot trajectory and its environment can be found in Fig.~\ref{fig:franka_traj_img}.

\newsavebox{\lam}

\begin{figure*}[t]
    \centering
    \savebox{\lam}{%
        \includegraphics[width=\textwidth]{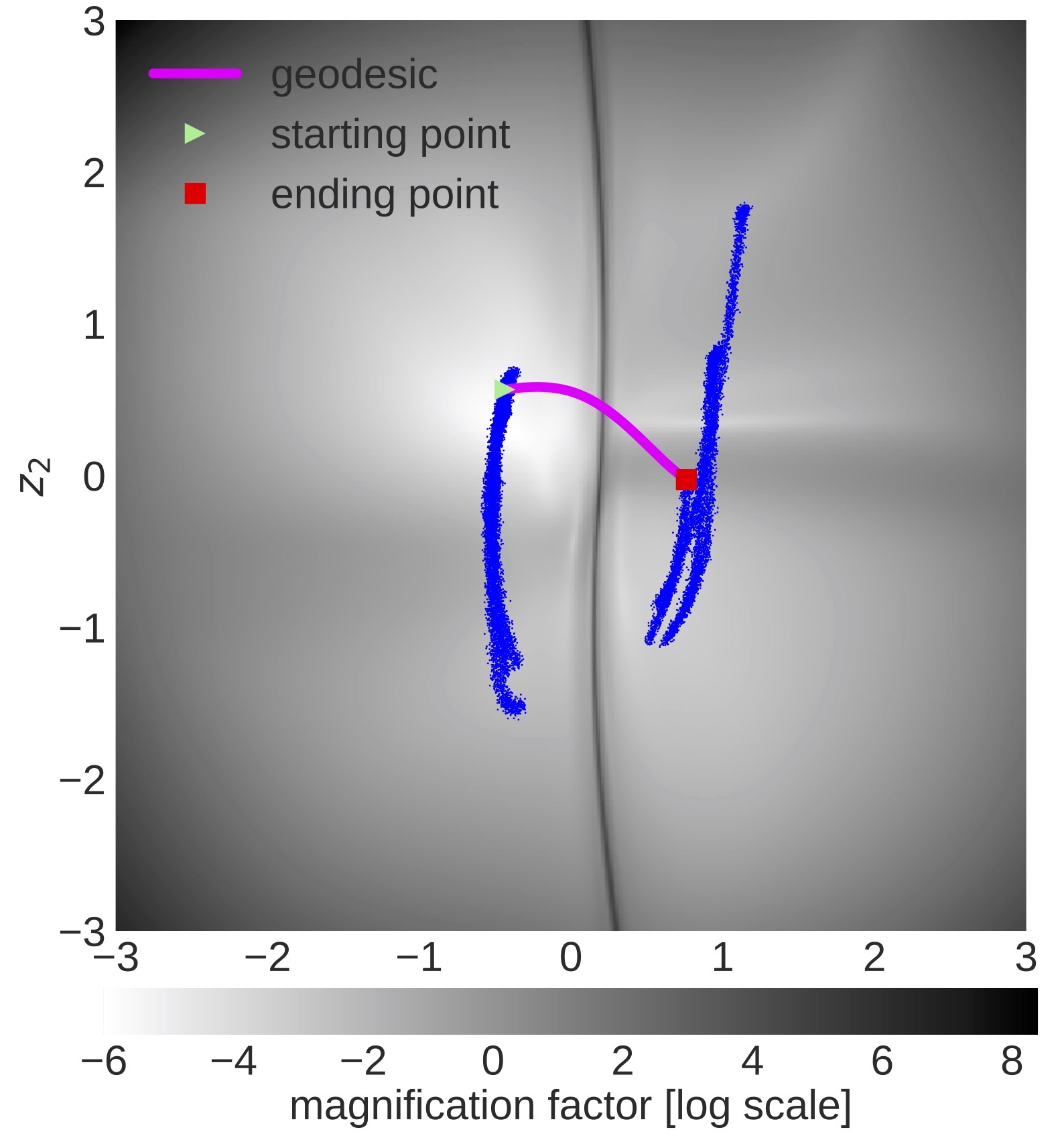}%
    }%
    \begin{subfigure}[b]{0.22\textwidth}
        \centering
        \includegraphics[width=\textwidth]{mf_franka_rotate_subset}
        \caption{Without active learning}%
        \label{fig:franka_latent_before}
    \end{subfigure}
    \hfill
    \begin{subfigure}[b]{0.22\textwidth}
        \centering
        \includegraphics[width=\textwidth]{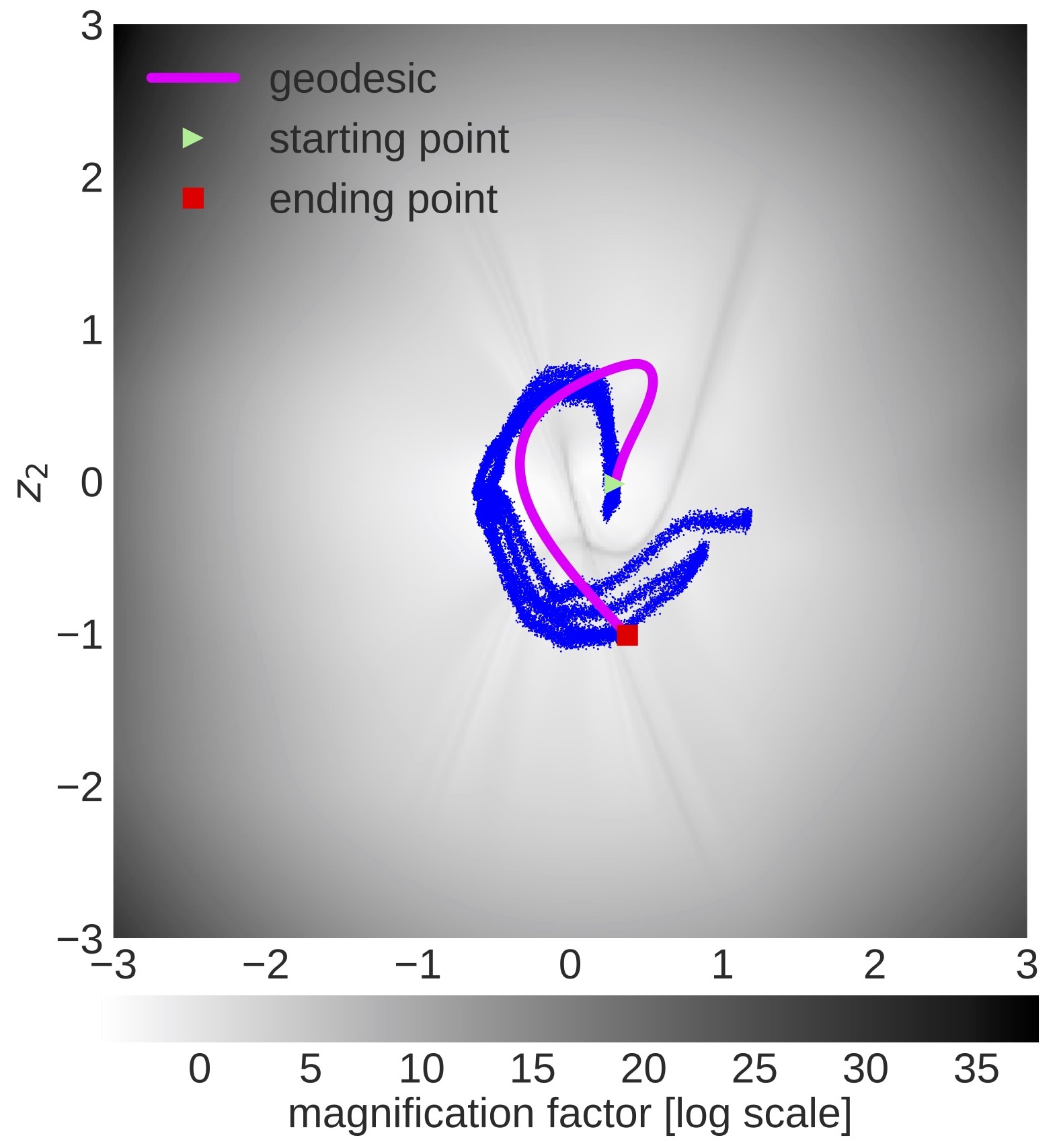}
        \caption{After active learning}%
        \label{fig:franka_latent_after}
    \end{subfigure}%
    \hfill
    \begin{subfigure}[b]{0.48\textwidth}
        \centering
		\includegraphics[width=\textwidth]{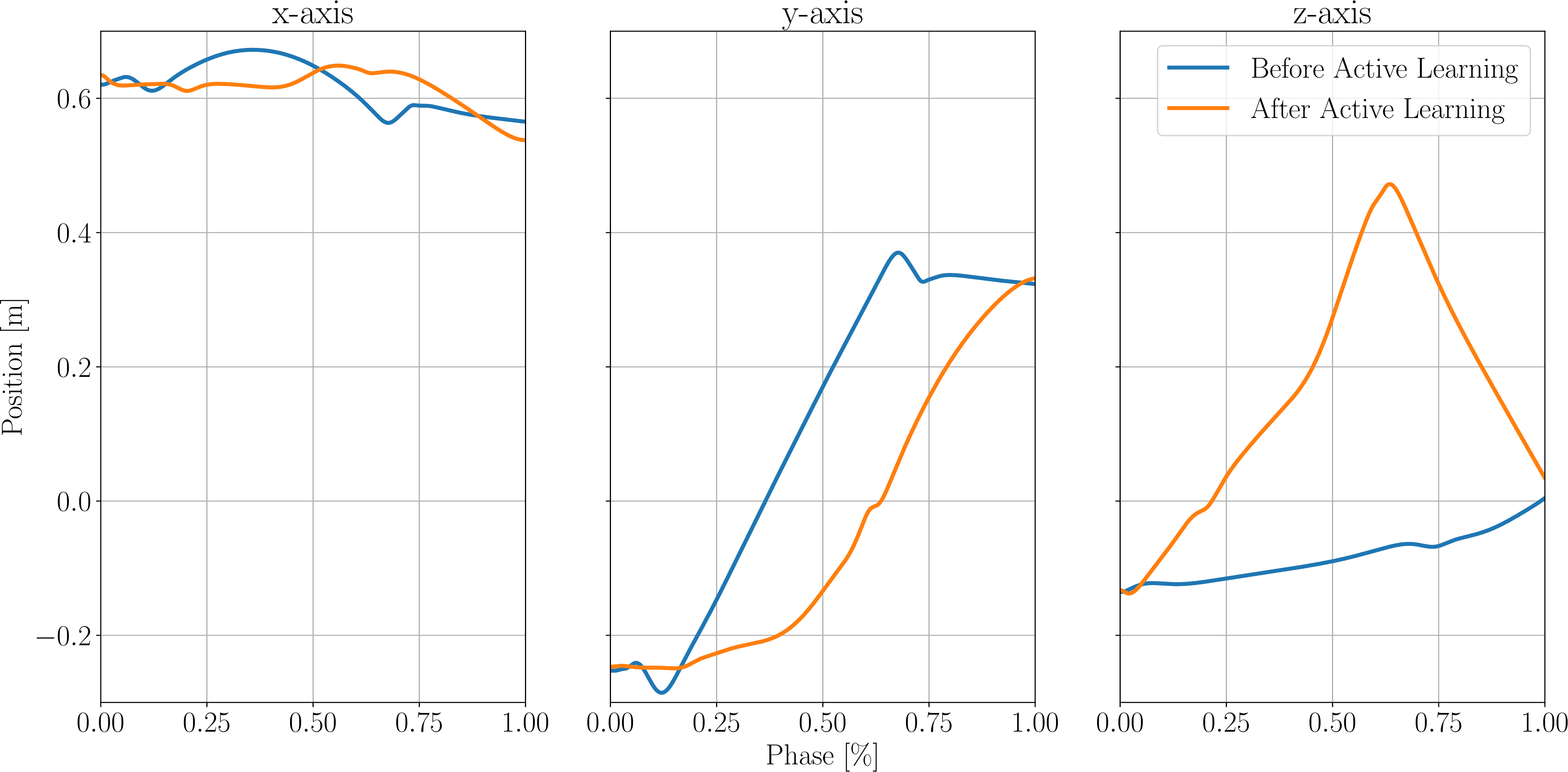}
        \caption{The Cartesian trajectories of the end-effector}%
        \label{fig:franka_traj}
    \end{subfigure}
    \caption{Trajectory generation for reaching movements while the robot avoids obstacles. 
	(\subref{fig:franka_latent_before}) The latent space after training with the initial demonstrations. 
	In blue we depict the training data points. 
	The generated trajectory crosses a high-$\MF$ region.
	(\subref{fig:franka_latent_after}) After providing the additional demonstration the resulting trajectory avoids collisions with the environment. 
	(\subref{fig:franka_traj}) Cartesian trajectories of the end-effector, before (orange) and after (blue) active learning. The trajectories after active learning are smoother and follow a path close to the demonstrations.}
\end{figure*}

%% file: sections/conclusion.tex
\section{Conclusion}
We introduced a new active-learning method, based on the model sensitivity in deep generative models. 
We showed that our method is suitable for efficiently learning new skills from demonstrations while maintaining some smoothness between known motions. 
In addition to triggering a query for new demonstrations, the magnification factor also indicates whether the observed data contains unrealistic postures, sudden fast movements, or indicates previously unseen/untrained movements.

Currently, the model is retrained when new data is acquired, in order to prevent the optimisation procedure to get stuck in local minima. We tackle this issue in future development and investigate alternative optimisation procedures that effectively allow for an online update.

%% file: sections/appendix.tex
\begin{table*}[!b]
    \caption{Parameters of the illustrative experiment}%
    \label{table:toy}\vspace{-1em}
    \begin{center}
        \begin{tabular}{lll}
            \toprule
            {\bf recognition model}  &{\bf generative model} &{\bf hyperparameters} \\
            \midrule
            Input $\in  \mathbb{R}^{2}$  &  Input $\in  \mathbb{R}^{2} $  &  learning rate = $2 \times 10^{-3}$ \\
            2 tanh FC $\times$ 512 units  & 2 tanh FC $\times$ 512 units & $K$ = 5\\
            linear FC output layer for means    &  softplus FC output layer for means & batch size = 150\\
            softplus FC output layer for variances  & RBF for variances &
            \\
            \bottomrule
        \end{tabular}
    \end{center}
\end{table*}

\begin{table*}[!b]
    \caption{Parameters for the pendulum experiment}%
    \label{table:pendulum}\vspace{-1em}
    \begin{center}
        \begin{tabular}{lll}
            \toprule
            {\bf recognition model}  &{\bf generative model} &{\bf hyperparameters} \\
            \midrule
            Input $\in  \mathbb{R}^{256}$  &  Input $\in  \mathbb{R}^{2} $  &  learning rate = $10^{-4}$ \\
            2 tanh FC $\times$ 512 units  & 10 residual $\times$ 128 units & $K$ = 5
            \\
            linear FC output layer for means    &  sigmoid FC output layer for means & batch size = 32\\
            softplus FC output layer for variances  & RBF for variances &\\
            \bottomrule
        \end{tabular}
    \end{center}
\end{table*}

\begin{table*}[!b]
    \caption{Parameters for the robot experiment}%
    \label{table:franka}\vspace{-1em}
    \begin{center}
        \begin{tabular}{lll}
            \toprule
            {\bf recognition model}  &{\bf generative model} &{\bf hyperparameters} \\
            \midrule
            Input $\in  \mathbb{R}^{7}$  &  Input $\in  \mathbb{R}^{2} $  &  learning rate = $5 \times10^{-4}$ \\
            2 tanh FC $\times$ 512 units  & 10 residual $\times$ 64 units & $K$ = 15\\
            linear FC output layer for means    &  softplus FC output layer for means & batch size = 150\\
            softplus FC output layer for variances  & RBF for variances &\\
            \bottomrule
        \end{tabular}
    \end{center}
\end{table*}

\balance%
\appendix

\section*{A. Details of the training procedure}\label{appendix:training_procedure}

The Adam optimiser~\cite{corrKingmaB14} was used for optimising the models of the three experiments. 
In the Tables~\ref{table:toy},~\ref{table:pendulum}, and~\ref{table:franka}, we
provide the parameters we used during training. We abbreviate the fully connected layers by FC.
Residual networks~\cite{he2016deep} are used for the pendulum and the FRANKA experiments. 
Increasing the depth of the generative model led to a more sensible and smoother magnification factor. 
With $K$, we refer to the number of importance-weighted samples.